\documentclass[runningheads]{llncs}
\usepackage[final]{changes}
\usepackage{times}
\usepackage{url}
\usepackage{latexsym}
\usepackage{graphicx}
\usepackage[ruled]{algorithm2e}
\usepackage{amsmath}
\usepackage[misc]{ifsym} 

\usepackage[colorlinks,
            linkcolor=black,
            anchorcolor=blue,
            citecolor=blue,
            urlcolor=blue
            ]{hyperref}
\begin{document}

\title{From Plots to Endings: A Reinforced Pointer Generator for Story Ending Generation}
%
%
\author{Yan Zhao\inst{1}\and
Lu Liu\inst{1}\and
Chunhua Liu\inst{1} \and Ruoyao Yang\inst{1} \and 
Dong Yu\inst{2,1}\Letter }
\authorrunning{Y. Zhao et al.}
%
\institute{Beijing Language and Culture University
\and
Beijing Advanced Innovation for Language Resources of BLCU
\email{\{zhaoyan.nlp,luliu.nlp,chunhualiu596\}@gmail.com}\\
\email{xmffaf@163.com}\\
\email{yudong\_blcu@126.com}}
\maketitle              
\begin{abstract}
\deleted{Story generation is a fundamental challenge in Natural Language Processing (NLP).}
We introduce a new task named Story Ending Generation (SEG), whic\-h aims at generating a coherent story ending from a sequence of story plot. We propose a framework consisting of a Generator and a Reward Manager for this task. The Generator follows the pointer-generator network with coverage mechanism to deal with out-of-vocabulary (OOV) and repetitive words. Moreover, a mixed loss method is introduced to enable the Generator to produce story endings of high semantic relevance with story plots. In the Reward Manager, the reward is computed to fine-tune the Generator with policy-gradient reinforcement learning (PGRL). We conduct experiments on the recently-introduced ROCStories Corpus. \deleted{To our knowledge, this is the first end-to-end model for SEG on this corpus.}We evaluate our model in both automatic evaluation and human evaluation. Experimental results show that our model exceeds the sequence-to-sequence baseline model by 15.75\% and 13.57\% in terms of CIDEr and consistency score respectively.

\end{abstract}

\section{Introduction}
\label{intro}

%
%
    %
    %
    %
    %
    %
    %

Story generation is an extremely challenging task in the field of NLP.
It has a long-standing tradition and many different systems have been proposed in order to solve the task. These systems are usually built on techniques such as planning~\cite{Meehan:1976:MWS:908045,DBLP:journals/corr/RiedlY14} and case-based reasoning~\cite{Gervás05storyplot,Stede:1996:SRT:974680.974687}, which rely on a fictional world including characters, objects, places, and actions. The whole system is very complicated and difficult to construct. 
\deleted{However, there are few research on generating a reasonable and sensible story ending from a sequence of story plot. }

We define a subtask of story generation named Story Ending Generation (SEG), which aims at generating a coherent story ending according to a sequence of story plot. A coherent ending should have a high correlation with the plot in terms of semantic relevance, consistency and readability. Humans can easily provide a logical ending according to a series of events in the story plot. The core objective of this task is to simulate the mode of people thinking to generate story endings, which has the significant application value in many artificial intelligence fields.

SEG can be considered as a Natural Language Generation (NLG) problem. Most studies on NLG aim at generating a target sequence that is semantically and lexically matched with the corresponding source sequence. Encoder-decoder framework for sequence-to-sequence learning~\cite{sutskever2014sequence} has been widely used in NLG tasks, such as machine translation~\cite{cho2014learning} and text summarization~\cite{chopra2016abstractive,nallapati2016abstractive,rush2015neural}.  Different from the above NLG tasks, SEG pays more attention to the consistency between story plots and endings. \deleted{Generally, a satisfying story ending is obtained via inference, that is, by finding the relevant parts in the corresponding story plot.} From different stories, we have observed that some OOV words in the plot, such as entities, may also appear in the ending. However, traditional sequence-to-sequence models replace OOV words by the special UNK token, which makes it unable to make correct predictions for these words. 
\deleted{Copy mechanism has been proved effective on this problem~\cite{gu2016incorporating,gulcehre2016pointing,miao2016language,see2017get,vinyals2015pointer}. }Moreover, encoder-decoder framework is of inability to avoid generating repetitive words.\deleted{~\cite{chen2016distraction,see2017get,tu2016modeling} apply coverage mechanism to handle repetition. } Another two limitations of encoder-decoder framework are exposure bias~\cite{ranzato2015sequence} and objective mismatch~\cite{phan2017consensus}, resulting from Maximum Likelihood Estimation (MLE) loss. To overcome these limitations, some methods~\cite{gu2016incorporating,liu2016improved,rennie2016self,see2017get,tu2016modeling} have been explored. 
 
\deleted{Recently, some research~\cite{liu2016improved,pasunuru2017reinforced,phan2017consensus,ranzato2015sequence,rennie2016self,wang2017video} has shown that both the two problems can be addressed by incorporating RL.} 

     
     
     
  
  
  


In this paper, we propose a new framework to solve the SEG problem. The framework consists of a Generator and a Reward Manager. The Generator follows a pointer-generator network to 
produce story endings. The Reward Manager is used for calculating the reward to fine tune the Generator through PGRL. With a stable and healthy environment that the Generator provides, PGRL can take effect to enable the generated story endings much more sensible.  
The key contributions of our model are as follows: 
\begin{itemize}
\item We apply copy and coverage mechanism~\cite{see2017get} to traditional sequence-to-sequence model as the Generator to handle OOV and repetitive words, improving the accuracy and fluency of generated story endings. 
\end{itemize}
\begin{itemize}
\item We add a semantic relevance loss to the original MLE loss as a new objective function to encourage the high semantic relevance between story plots and generated endings.
\end{itemize}
\begin{itemize}
\item We define a Reward Manager to fine tune the Generator through PGRL. In the Reward Manager, we attempt to use different evaluation metrics as reward functions to simulate the process of people writing a story.
\end{itemize}

We conduct experiments on the recently-introduced ROCStories Corpus~\cite{mostafazadeh2016corpus}.\deleted{(as shown in Table~\ref{story_eg}). To our knowledge, this is the first end-to-end model for SEG on this corpus.} We utilize both automatic evaluation and human evaluation to evaluate our model. There are word-overlap and embedding metrics in the automatic evaluation~\cite{sharma2017relevance}.
In the human evaluation, we evaluate generated endings in terms of consistency and readability, which reflect the logical coherence and fluency of those endings.
Experimental results demonstrate that our model outperforms previous basic neural generation models in both automatic evaluation and human evaluation. Better performance in consistency indicates that our model has strong capability to produce reasonable sentences.

\section{Related Work}
\deleted{SEG is a new task so that there is no work specifically on it. In this section, we focus on Neural Network (NN) models for some specific NLG tasks.}

\subsection{Encoder-decoder Framework} Encoder-decoder framework, which uses neural networks as encoder and decoder, was first proposed in machine translation~\cite{cho2014learning,sutskever2014sequence} and has been widely used in NLG tasks. The encoder reads and encodes a source sentence into a fixed-length vector, then the decoder outputs a new sequence from the encoded vector. Attention mechanism~\cite{bahdanau2014neural} extends the basic encoder-decoder framework by assigning different weights to input words when generating each target word.~\cite{chopra2016abstractive},~\cite{nallapati2016abstractive} and~\cite{rush2015neural} apply attention-based encoder-decoder model to text summarization.\deleted{~\cite{hermann2015teaching} and~\cite{kadlec2016text} develop new and effective attentive models for machine reading comprehension.}

\subsection{Copy and Coverage Mechanisms} The encoder-decoder framework is unable to deal with OOV words. In most NLP systems, there usually exists a predefined vocabulary, which only contains top-K most frequent words in the training corpus. All other words are called OOV and replaced by the special UNK token. This makes neural networks difficult to learn a good representation for OOV words and some important information would be lost. To tackle this problem,~\cite{vinyals2015pointer} and~\cite{gulcehre2016pointing} introduce pointer mechanism to predict the output words directly from the input sequence.~\cite{gu2016incorporating} incorporate copy mechanism into sequence-to-sequence models and propose CopyNet to naturally combine generating and copying. Other extensions of copy mechanism appear successively, such as ~\cite{miao2016language}. Another problem of the encoder-decoder framework is repetitive words in the generated sequence. Accordingly coverage model in~\cite{tu2016modeling} maintains a coverage vector for keeping track of the attention history to adjust future attention.\deleted{~\cite{chen2016distraction} also utilize the coverage mechanism called distraction for summarization.} A hybrid pointer-generator network introduced by~\cite{see2017get} combines copy and coverage mechanism to solve the above problems.

\subsection{Reinforcement Learning for NLG} The encoder-decoder framework is typically trained by maximizing the log-likelihood of the next
word given the previous ground-truth input words, resulting in exposure bias~\cite{ranzato2015sequence} and objective mismatch~\cite{phan2017consensus} problems. Exposure bias refers to the input distribution discrepancy between training and testing time, which makes generation brittle as error accumulate. Objective mismatch refers to using MLE at training time while using discrete and non-differentiable NLP metrics such as BLEU at test time. Recently, it has been shown that both the two problems can be addressed by incorporating RL in captioning tasks. Specifically,~\cite{ranzato2015sequence} propose the MIXER algorithm to directly optimize the sequence-based test metrics.~\cite{liu2016improved} improve the MIXER algorithm and uses a policy gradient method.~\cite{rennie2016self} present a new optimization approach called self-critical sequence training (SCST). Similar to the above methods,~\cite{pasunuru2017reinforced},~\cite{phan2017consensus} and~\cite{wang2017video} explore different reward functions for video captioning. Researchers also make attempts on other NLG tasks such as dialogue generation~\cite{li2016deep}, sentence simplification~\cite{zhang2017sentence} and abstract summarization~\cite{paulus2017deep}, obtaining satisfying performances with RL.

Although many approaches for NLG have been proposed, SEG is still a challenging yet interesting task and worth trying.

\section{\label{models} Models}
\begin{figure*}[t]
	\centering
  \includegraphics[width=\textwidth]{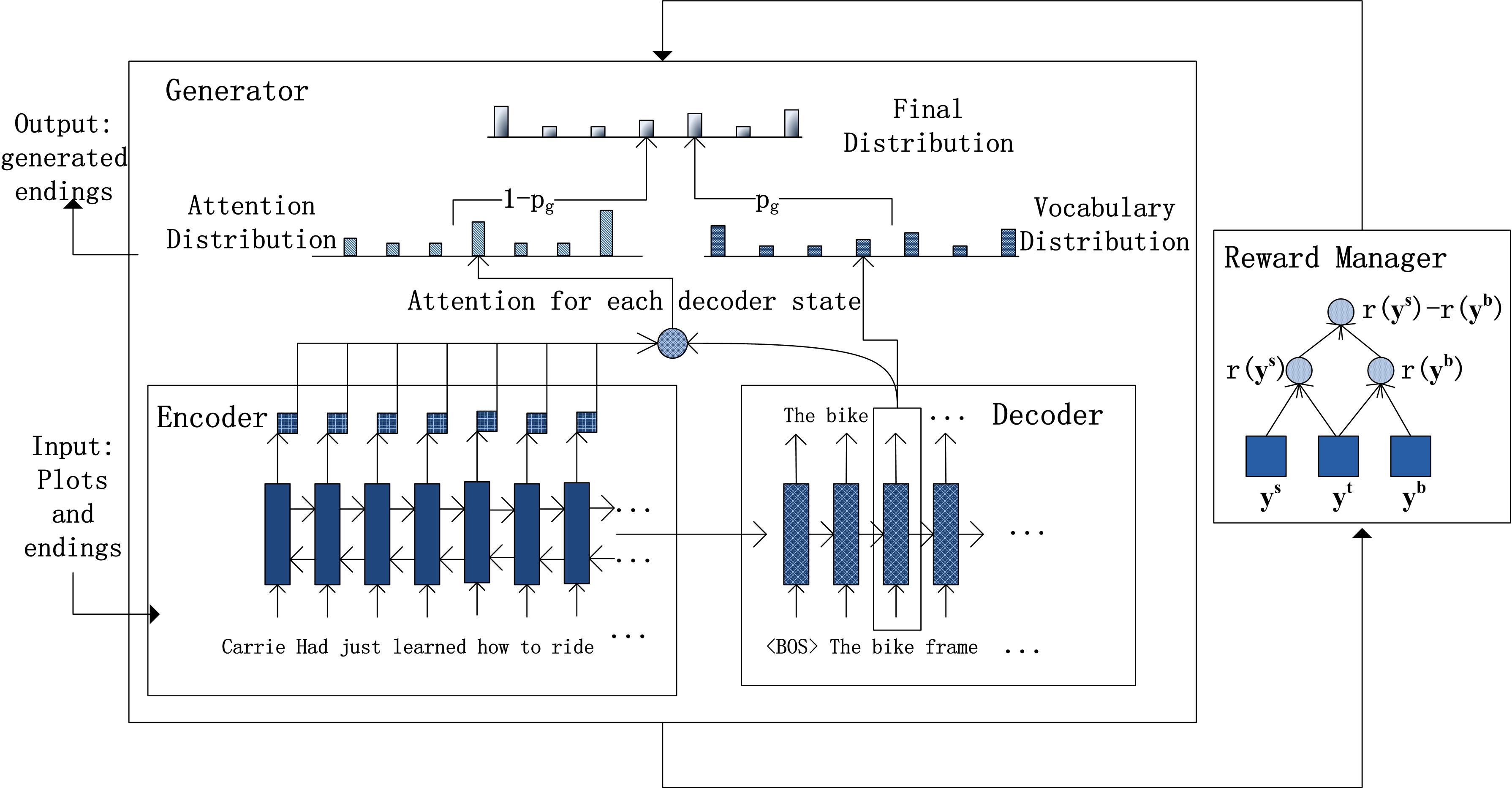}
	\caption{Overview of our model}
	\label{fig:model}
\end{figure*}

Figure~\ref{fig:model} gives the overview of our model. It contains a Generator and a Reward Manager. The Generator follows the pointer-generator network with coverage mechanism to address the issues of OOV words and repetition. A mixed loss method is exploited in the Generator for improving semantic relevance between story plots and generated endings. The Reward Manager is utilized to produce the reward for PGRL. The reward can be calculated by evaluation metrics or other models in the Reward Manager. Then it is passed back to the Generator for updating parameters.
Following sections give more detailed descriptions of our models.

\subsection{\label{attn} Attention-based Encoder-decoder Model}
Our attention-based encoder-decoder baseline model is similar to the framework in~\cite{bahdanau2014neural}. Given a sequence of plot words of length \(T_e\), we feed the word embeddings into a single-layer bidirectional LSTM to compute a sequence of encoder hidden states \(h_i^e = \{h_1^e,h_2^e,...,h_{T_e}^e\}\). At each decoding step {\em t}, a single LSTM decoder takes the previous word embedding and context vector \(c_{t-1}\), which is calculated by attention mechanism, as inputs to produce decoder hidden state \(h_t^d\). 

We concatenate the context vector \(c_t\) and decoder hidden state \(h_t^d\) to predict the probability distribution \(P_{v}\) over all the words in the vocabulary:
\begin{equation}\begin{split} P_{v} = softmax(W_1(W_2[h_t^d,c_t] + b_2) + b_1)\end{split}\end{equation} where \(W_1\), \(W_2\), \(b_1\), \(b_2\) are all learnable parameters. [a,b] means the concatenation of a and b.

MLE is usually used as the training objective for sequence-to-sequence tasks. We denote \(y_t^* = \{y_1^*,y_2^*,...,y_{T_d}^*\}\) as the ground truth output ending. The cross entropy loss function is defined as:
\begin{equation} \label{L_mle} L_{mle} = -\sum_{t=1}^{T_d} \log P_{v}(y_t^*)\end{equation}

\subsection{Pointer-generator Network with Coverage Mechanism}
From the dataset, we find that some words in the story plot will also appear in the ending. It makes sense that the story ending usually describes the final states of some entities, which are related to the events in the story plot.
Thus we follow the hybrid pointer-generator network in~\cite{see2017get} to copy words from the source plot text via pointing~\cite{vinyals2015pointer}, in this way we can handle some OOV words. 
In this model, we accomplish the attention-based encoder-decoder model in the same way as Section~\ref{attn}. Additionally, we choose top-k words to build a vocabulary and calculate a generation probability \(p_{g}\) to weight the probability of generating words from the vocabulary.
\begin{equation} p_{g} = sigmoid (W_{c_t}c_t + W_{h_t}h_t + W_{y_t}y_t + b_p)\end{equation} 
where \(c_t\), \(h_t^d\), \(y_t\) represent the context vector, the decoder hidden state and the decoder input at each decoding step {\em t} respectively. \(W_{c_t}\), \(W_{h_t}\), \(W_{y_t}\)  are all weight parameters and \(b_p\) is a bias. 

Furthermore, the attention distributions of duplicate words are merged as \(P_{att}(w_t)\).
We compute the weighted sum of vocabulary distribution \(P_{v}(w_t)\) and \(P_{att}(w_t)\) as the final distribution:
\deleted{\begin{equation} P_{att}(w_t) =\sum_{i}^{w_t=w_i}\alpha_i^t\end{equation}}
\begin{equation} P_{fin}(w_t) = p_{g}P_{v}(w_t) + (1 - p_{g})P_{att}(w_t)\end{equation}  

The loss function is the same as that in attention-based encoder-decoder model, with \(P_{v}(w_t)\) in equation (\ref{L_mle}) changed to \(P_{fin}(w_t)\).

\deleted{The coverage vector stands for the degree of coverage that words in source plot text have received from the attention mechanism. Inspired by ~\cite{see2017get},}

To avoid repetition, we also apply coverage mechanism~\cite{tu2016modeling} to track and control coverage of the source plot text.
We utilize the sum of attention distributions over all previous decoder steps as  the coverage vector \(s^t\). Then the coverage vector is added into the calculation of attention score \(e_i^t\) to avoid generating repetitive words:
\begin{equation} e_i^t = v^Ttanh(W_1^{att}h_i^e + W_2^{att}h_t^d  + W_3^{att}s_i^t)\end{equation} where \(W_1^{att}\), $W_2^{att}$, $W_3^{att}$, and $v^T$ are learnable parameters. 

\deleted{, formulated as follows:
\begin{equation} cov^t = \sum_{t'=0}^{t-1} \alpha^{t'}\end{equation}
Note that \(cov^0\) is a zero vector.}

Moreover, a coverage loss is defined and added to the loss function to penalize repeatedly attending to the same locations:
\begin{equation}L_{poi} = -\sum_{t=1}^{T_d} [\log P_{fin}(w_t) \\+ \beta \sum_{i=1}^{T_e} min(\alpha_i^t, s_i^t)]\end{equation} 
where \(\beta\) is a hyperparameter and {\em min(a,b)} means the minimum of a and b. 

\subsection{\label{mix_loss} Mixed Loss Method}

Pointer-generator network has the capacity of generating grammatically and lexically accurate story endings. These story endings are usually of low semantic relevance with plots, which fails to meet our requirements of satisfying story endings. To overcome this weakness, we add a semantic similarity loss to the original loss as the new objective function. 

There are some different ways to obtain the semantic vectors, such as the average pooling of all word embeddings or max pooling of the RNN hidden outputs. Intuitively, the bidirectional LSTM encoder can fully integrate the context information from two directions. Therefore the last hidden output of the encoder \(h_{T_e}^e\) is qualified to represent the semantic vector of the story plot. Similar to ~\cite{ma2017semantic}, we select \(h_{T_e}^e\) as the plot semantic vector \(v_{plot}\), and the last hidden output of decoder subtracting last hidden output of the encoder as the semantic vector of the generated ending \(v_{gen}\) :
\begin{gather}v_{plot}=h_{T_e}^e\\
v_{gen}=h_{T_d}^d-h_{T_e}^e \end{gather}

{\bf Semantic Relevance:} Cosine similarity is typically used to measure the matching affinity between two vectors. With the plot semantic vector \(v_{plot}\) and the generated semantic vector \(v_{gen}\), the semantic relevance is calculated as:\begin{equation} S_{sem}= cos(v_{plot},v_{gen})= \frac{v_{plot} \cdot v_{gen}}{\|v_{plot}\|\|v_{gen}\|} \end{equation}

{\bf Mixed Loss:} Our objective is maximizing the semantic relevance between story plots and generated endings. As a result, we combine the similarity score \(S_{sem}\) with the original loss as a mixed loss:
\begin{equation} L_{mix}= -S_{sem} + L_{poi}\end{equation}

The mixed loss method encourages our model to generate story endings of high semantic relevance with plots. In addition, it makes the Generator more stable for applying RL algorithm.

%
\subsection{Policy-gradient Reinforcement Learning}
The Generator can generate syntactically and semantically correct sentences with the above two methods. However, models trained with MLE still suffer from exposure bias~\cite{ranzato2015sequence} and objective mismatch~\cite{phan2017consensus} problems. 
A well-known policy-gradient reinforcement learning algorithm~\cite{williams1992simple} can directly optimize the non-differentiable evaluation metrics such as BLEU, ROUGE and CIDEr. It has good performance on several sequence generation tasks~\cite{paulus2017deep,rennie2016self}.

In order to solve the problems, we cast the SEG task to the reinforcement learning framework. An \textit{agent} interacting with the external environment in reinforcement learning can be analogous to our generator taking words of the story plot as inputs and then producing outputs. The parameters of the agent define a \textit{policy}, which results in the agent picking an \textit{action}. In our SEG task, an action refers to generating a sequence as story ending. After taking an action,  the agent computes the \textit{reward} of this action and updates its internal \textit{state}.

\begin{algorithm*}[t]
\label{algo}
\caption{The reinforcement learning algorithm for training the Generator $G_{{\theta}'}$ }
\LinesNumbered 
\KwIn{ROCstories \{($x,y$)\};}
\KwOut{Generator $G_{{\theta}'}$;}
Initialize $G_{\theta}$ with random weights $\theta$\; 
Pre-train $G_{\theta}$ using MLE on dataset \{($x,y$)\}\;
Initialize $G_{{\theta}'}$ = $G_{\theta}$\; 
\For{each epoch}{
	Generate an ending $y^b=(y_1^b,\ldots, y_T^b)$ according to $G_{{\theta}'}$ given $x$\;
    Sample an ending $y^s= (y_1^s,\ldots,y_{T}^s)$ from the probability distribution $P(y_t^s)$\;
  	Compute reward $r(y^b)$ and $r(y^s)$ defined in the Reward Manager\;
    Compute $L_{rl}$ using Eq.(\ref{l_rl})\;
    Compute $L_{total}$ using Eq.(\ref{l_total})\;
    Back-propagate to compute $\nabla _ {{\theta}'} \mathcal{L}_{total}({\theta}')$\;
    Update Generator $G_{{\theta}'}$ using ADAM optimizer with learning rate $lr$       
}
\Return $G_{{\theta}'}$
\end{algorithm*}

Particularly, we use the SCST approach~\cite{rennie2016self} to fine-tune the Generator. This approach designs a loss function, which is formulated as :
 \begin{equation} \label{l_rl} L_{rl} = (r(y^b)-r(y^s))\sum_{t=1}^{T}\log P(y_t^s) \end{equation}
where \(y^s = (y_1^s,...,y_{T}^s)\) is a sequence sampled from the probability distribution \(P(y_t^s)\) at each decoding time step {\em t}. \(y^b\) is the baseline sequence obtained by greedy search from the current model. \(r(y)\) means the reward for the sequence \(y\), computed by the evaluation metrics. Intuitively, the loss function \(L_{rl}\) enlarges the log-probability of the sampled sequence \(y^s\) if it obtains a higher reward than the baseline sequence \(y^b\). In the Reward Manager, we try several different metrics as reward functions and find that BLEU-4 produces better results than others.

To ensure the readability and fluency of the generated story endings, we also define a blended loss function, which is a weighted combination of the mixed loss in Section~\ref{mix_loss} and the reinforcement learning loss:
\begin{equation} \label{l_total} L_{total}=\mu L_{rl} +(1-\mu) L_{mix}\end{equation}
where $\mu$ is a hyper-parameter controlling the ratio of $ L_{rl}$ and $L_{mix}$. This loss function can make a trade-off between the RL loss and mixed loss in Section ~\ref{mix_loss}.

The whole reinforcement learning algorithm for training the Generator is summarized as Algorithm~\ref{algo}. \deleted{We first pre-train the Generator to create a appropriate environment for RL. Then we use the SCST approach to update the parameters of the Generator by policy gradient. The improved Generator is capable of generating satisfying story endings.}
\setcounter{footnote}{0}
\section{Experiments}

\subsection{Dataset}
\textbf{ROCStories Corpus} is a publicly available collection of short stories released by~\cite{mostafazadeh2016corpus}. There are 98161 stories in training set and 1871 stories in both validation set and test set. A complete story in the corpus consists of five sentences, in which the first four and last one are viewed as the plot and ending respectively. The corpus captures a variety of causal and temporal commonsense relations between everyday events. We choose it for our SEG task because of its great performance in quantity and quality.

\subsection{Experimental Setting}

In this paper, we choose attention-based sequence-to-sequence model (Seq2Seq) as our baseline. Additionally, we utilize pointer-generator network with coverage mechanism (PGN) to deal with OOV words and avoid repetition. Then we train pointer-generator network with mixed loss method (PGN+Sem\_L) and PGRL algorithm (PGN+RL) respectively. Finally, we integrate the entire model with both mixed loss method and PGRL algorithm (PGN+Sem\_L+RL).

We implement all these models with Tensorflow~\cite{DBLP:journals/corr/AbadiBCCDDDGIIK16}. In all the models, the LSTM hidden units, embedding dimension, batch size, dropout rate and beam size in beam search decoding are set to 256, 512, 64, 0.5 and 4 respectively. We use ADAM~\cite{kingma2014adam} optimizer with an initial learning rate of 0.001 when pre-training the generator and $5 \times 10^{-5}$ when running RL training. The weight coefficient of coverage loss \(\beta\) is set to 1. The ratio \(\mu\) between RL loss and mixed loss is 0.95. Through counting all the words in the training set, we obtain the vocab size 38920 (including extra special tokens UNK, PAD and BOS). The size of vocabulary is 15000 when training the pointer-generator network. The coverage mechanism is used after 10-epoch training of single pointer-generator network. We evaluate the model every 100 global steps and adopt early stopping on the validation set.

\subsection{Evaluation Metrics}
\deleted{It is very difficult to choose appropriate evaluation metrics for NLG tasks. }For SEG, a story may have different kinds of appropriate endings for the same plot. It is unwise to evaluate the generated endings from a single aspect. Therefore we apply automatic evaluation and human evaluation in our experiments.

\textbf{Automatic Evaluation:} We use the evaluation package nlg-eval\footnote{https://github.com/Maluuba/nlg-eval}~\cite{sharma2017relevance}, which is a publicly available tool supporting various unsupervised automated metrics for NLG. It considers not only word-overlap metrics such as BLEU, METEOR, CIDEr and ROUGE, but also embedding-based metrics including SkipThoughts Cosine Similarity (STCS), Embedding Average Cosine Similarity (EACS), Vector Extrema Cosine Similarity (VE\-CS), and Greedy Matching Score (GMS). 

\textbf{Human Evaluation:} We randomly select 100 stories from test set and \deleted{generate story endings using models described in Section~\ref{models}.We}define two criteria to implement human evaluation. Consistency refers to the logical coherence and accordance between story plots and endings. Readability measures the quality of endings in grammar and fluency. Five human assessors are asked to rate the endings on a scale of 0 to 5.

\begin{table*}[t]
\begin{center}
\begin{tabular}{cccccccc}
\hline

\bf Models& \bf BLEU-1&\bf BLEU-2& \bf BLEU-3& \bf  BLEU-4& \bf METEOR& \bf ROUGE-L& \bf CIDEr\\
\hline
Seq2Seq& 26.17& 10.54 &5.29&3.03&10.80&26.84&47.48\\
PGN& 28.07& 11.39& 5.53& 3.02& 10.87& 27.80&  51.09\\
PGN+Sem\_L& 28.21& 11.56& 5.81& 3.33& 11.08& 28.15& 53.21\\
PGN+RL& 28.05& 11.50& 5.69& 3.17& 10.83& 27.46& 49.83\\
PGN+Sem\_L+RL&\bf 28.51& \bf 11.92& \bf 6.16& \bf 3.53& \bf 11.10& \bf 28.52& \bf 54.96\\
\hline
\end{tabular}
\end{center}
\caption{\label{me_wo}Results on Word-overlap Metrics.}
\end{table*}

\begin{table*}[t]
\begin{center}
\begin{tabular}{ccccccccc}
\hline
\bf Models& \bf STCS-p& \bf EACS-p& \bf VECS-p& \bf GMS-p\\
\hline
Ground Truth & 66.94&  87.03&  45.64&  70.75\\
Seq2Seq& 67.94&	89.98&	46.37&	73.23\\
PGN&68.15& 89.20& 48.96& 74.64\\
PGN+Sem\_L&  67.90& 90.02& 48.60& 74.61\\
PGN+RL&  68.07& 89.97& 49.48& 74.90\\
PGN+Sem\_L+RL& 67.84& 89.50& 48.44& 74.45\\
\hline
\end{tabular}
\end{center}
\caption{\label{me_e} Results on Embedding Based Metrics.}
\end{table*}
\subsection{Automatic Evaluation}

\subsubsection{Results on Word-overlap Metrics}

Results on word-overlap metrics are shown in Table~\ref{me_wo}. Obviously, PGN+sem\_L+RL achieves the best result among all the models. \deleted{It exceeds the baseline model (Seq2Seq) by (2.34 BLEU-1, 1.38 BLEU-2, 0.87 BLEU-3, 0.50 BLEU-4, 0.30 METEOR, 1.68 ROUGE-L, 7.48 CIDEr). }This indicates that our methods are effective on producing accurate story endings.

From the results, we have some other observations. PGN surpasses the Seq2Seq baseline model, especially in BLEU-1 (+1.9) and CIDEr (+3.61). This behaviour suggests that copy and coverage mechanisms can effectively handle OOV and repetitive words so as to improve scores of word-overlap metrics. Compared with PGN, the results of PGN+Sem\_L have an increase in all the word-overlap metrics. This improvement benefits from our mixed loss method based on semantic relevance. More interestingly, PGN+RL performs poorly while PGN+Sem\_L+RL obtains an improvement. We attribute this to an insufficiency of applying RL directly into PGN. \deleted{RL has an advantage in fine-tuning the pre-trained model which is in a basically stable environment. The environment PGN+Sem\_L created is healthy and qualified enough for incorporating RL.} Results on PGN+Sem\_L+RL prove that mixed loss method shows its effectiveness and it motivates RL to take effect.

\subsubsection{Results on Embedding Based Metrics}

We compute cosine similarities between generated endings and plots. For comparison, the cosine similarity between target endings and plots is provided as the ground-truth reference. Evaluation results are illustrated in Table~\ref{me_e}.

\begin{table*}[t]
\begin{center}
\begin{tabular}{cccc}
\hline
\bf Models&\bf  Consistency&\bf   Readability\\
\hline
Ground Truth& 4.33&  4.83\\
Seq2Seq& 2.80&		4.33\\
PGN&2.95& 4.38\\
PGN+Sem\_L&  3.00& \bf 4.43\\
PGN+RL& 2.92&  4.36\\
PGN+Sem\_L+RL& \bf 3.18& 4.41\\
\hline
\end{tabular}
\end{center}
\caption{\label{me_h} Human Evaluation Results.}
\end{table*}
\begin{table*}[t]
\begin{center}
    \begin{tabular}{| p{2.7cm} | p{4.5cm} |p{4.5cm} |}
    \hline \bf Model & \bf Example-1& \bf Example-2\\
    \hline
		Plot &  Juanita realizes that she needs warmer clothing to get through winter. She looks for a jacket but at first everything she finds is expensive. Finally she finds a jacket she can afford. She buys the jacket and feels much better.& My dad took me to a baseball game when I was little. He spent that night teaching me all about the sport. He showed me every position and what everything meant. He introduced me to one of my favorite games ever.\\ 
	\hline
       	Target&  She is happy.&  Now, playing or seeing baseball on TV reminds me of my father.\\
    \hline
    	 Seq2Seq& Juanita is happy that she is happy that she is happy. & I was so happy that he was so happy.\\
    \hline
        	PGN& Juanita is happy that she needs through winter clothing.& I was so excited to have a good time.\\
    \hline
        	PGN+Sem\_L& Juanita is happy to have warmer clothing to winter.& I was so happy to have a good time.\\
    \hline
        	PGN+RL& Juanita is happy that she has done through winter.& My dad told me I had a great time.\\
    \hline
        	PGN+Sem\_L+RL& Juanita is happy that she has \textbf{a new warmer clothing}.& I am going to \textbf{play with} my dad.\\
    \hline
        
    \end{tabular}
\end{center}
\caption{\label{story_g} Examples of plots, target endings and generated endings of all models}

\end{table*}

Embedding-based metrics tend to acquire more semantics than word-overlap metrics. 
From Table~\ref{me_e}, all the models are likely to generate endings with less discrepancy in terms of the embedding based metric. It can also be observed that scores of all models surpass that of the ground-truth reference. This indicates that nearly every model can generate endings which have higher cosine similarity scores with the plot. But it cannot just measure these endings by calculating these scores.

\subsection{Human Evaluation}
\deleted{We randomly select 100 stories from test set and evaluate endings generated by different models.}Table ~\ref{me_h} presents human evaluation results. Apparently, PGN+Sem\_L+RL and PGN+Se\-m\_L achieve the best results in terms of consistency and readability respectively. The readability score of PGN+sem\_L+RL is good enough, with the difference of 0.02 compared to the best result (PGN+Sem\_L). We can also observe that readability scores of all the models are basically equivalent. It manifests that all the models have the ability to generate grammatically and lexically correct endings. Therefore, we only analyze the consistency scores as follows.

The consistency score of PGN increases by 5.37\% compared with Seq2Seq. This is attributed to the copy and coverage mechanism discouraging OOV and repetitive words. The score of PGN+Sem\_L is 1.69\% higher than PGN. With mixed loss method, the semantic relevance between story plots and endings is improved, leading to better performance in consistency. PGN+RL gets a lower score than PGN. This indicates that PGN is not prepared for incorporating RL, and RL alone can not directly promote PGN. In contrast, the score of PGN+Sem\_L+RL is 6\% higher than PGN+Sem\_L. We can conclude that PGN with mixed loss method rather than simple PGN is more capable of stimulating RL to take effect.

In order to demonstrate the generative capability of different models, we present some endings generated by different models in Table~\ref{story_g}. Compared with other models, the endings generated by PGN+Sem\_L+RL are not only fluent but also contain new information (words that are bold). Thus, we conclude that our model reaches its full potential under the joint of mixed loss method and RL.

\section{Conclusion}

In this work we propose a framework consisting of a Generator and a Reward Manager to solve the SEG problem. Following the pointer-generator network with coverage mechanism, the Generator is capable of handling OOV and repetitive words. A mixed loss method is also introduced to encourage the Generator to produce story endings of high semantic relevance with story plots. The Reward Manager can fine tune the Generator through policy-gradient reinforcement learning, promoting the effectiveness of the Generator.
Experimental results on ROCStories Corpus demonstrate that our model has good performance in both automatic evaluation and human evaluation.  

\section*{Acknowledgements}
This work is funded by Beijing Advanced Innovation for Language Resources of BLCU, the Fundamental Research Funds for the Central Universities in BLCU (No. 17PT05).


\bibliographystyle{splncs04}

\bibliography{seg}

\begin{thebibliography}{10}
\providecommand{\url}[1]{\texttt{#1}}
\providecommand{\urlprefix}{URL }
\providecommand{\doi}[1]{https://doi.org/#1}

\bibitem{DBLP:journals/corr/AbadiBCCDDDGIIK16}
Abadi, M., Barham, P., Chen, J., Chen, Z., Davis, A., Dean, J., Devin, M.,
  Ghemawat, S., Irving, G., Isard, M., Kudlur, M., Levenberg, J., Monga, R.,
  Moore, S., Murray, D.G., Steiner, B., Tucker, P.A., Vasudevan, V., Warden,
  P., Wicke, M., Yu, Y., Zhang, X.: Tensorflow: {A} system for large-scale
  machine learning. CoRR  \textbf{abs/1605.08695} (2016),
  \url{http://arxiv.org/abs/1605.08695}

\bibitem{bahdanau2014neural}
Bahdanau, D., Cho, K., Bengio, Y.: Neural machine translation by jointly
  learning to align and translate. arXiv preprint arXiv:1409.0473  (2014)

\bibitem{cho2014learning}
Cho, K., Van~Merri{\"e}nboer, B., Gulcehre, C., Bahdanau, D., Bougares, F.,
  Schwenk, H., Bengio, Y.: Learning phrase representations using rnn
  encoder-decoder for statistical machine translation. arXiv preprint
  arXiv:1406.1078  (2014)

\bibitem{chopra2016abstractive}
Chopra, S., Auli, M., Rush, A.M.: Abstractive sentence summarization with
  attentive recurrent neural networks. In: Proceedings of the 2016 Conference
  of the North American Chapter of the Association for Computational
  Linguistics: Human Language Technologies. pp. 93--98 (2016)

\bibitem{Gervás05storyplot}
Gervás, P., Díaz-agudo, B., Peinado, F., Hervás, R.: Story plot generation
  based on cbr. Journal of Knowledge Based Systems  \textbf{18}, ~2--3 (2005)

\bibitem{gu2016incorporating}
Gu, J., Lu, Z., Li, H., Li, V.O.: Incorporating copying mechanism in
  sequence-to-sequence learning. arXiv preprint arXiv:1603.06393  (2016)

\bibitem{gulcehre2016pointing}
Gulcehre, C., Ahn, S., Nallapati, R., Zhou, B., Bengio, Y.: Pointing the
  unknown words. arXiv preprint arXiv:1603.08148  (2016)

\bibitem{kingma2014adam}
Kingma, D.P., Ba, J.: Adam: A method for stochastic optimization. arXiv
  preprint arXiv:1412.6980  (2014)

\bibitem{li2016deep}
Li, J., Monroe, W., Ritter, A., Galley, M., Gao, J., Jurafsky, D.: Deep
  reinforcement learning for dialogue generation. arXiv preprint
  arXiv:1606.01541  (2016)

\bibitem{liu2016improved}
Liu, S., Zhu, Z., Ye, N., Guadarrama, S., Murphy, K.: Improved image captioning
  via policy gradient optimization of spider. arXiv preprint arXiv:1612.00370
  (2016)

\bibitem{ma2017semantic}
Ma, S., Sun, X.: A semantic relevance based neural network for text
  summarization and text simplification. arXiv preprint arXiv:1710.02318
  (2017)

\bibitem{Meehan:1976:MWS:908045}
Meehan, J.R.: The Metanovel: Writing Stories by Computer. Ph.D. thesis, New
  Haven, CT, USA (1976), aAI7713224

\bibitem{miao2016language}
Miao, Y., Blunsom, P.: Language as a latent variable: Discrete generative
  models for sentence compression. arXiv preprint arXiv:1609.07317  (2016)

\bibitem{mostafazadeh2016corpus}
Mostafazadeh, N., Chambers, N., He, X., Parikh, D., Batra, D., Vanderwende, L.,
  Kohli, P., Allen, J.: A corpus and evaluation framework for deeper
  understanding of commonsense stories. arXiv preprint arXiv:1604.01696  (2016)

\bibitem{nallapati2016abstractive}
Nallapati, R., Zhou, B., Gulcehre, C., Xiang, B., et~al.: Abstractive text
  summarization using sequence-to-sequence rnns and beyond. arXiv preprint
  arXiv:1602.06023  (2016)

\bibitem{pasunuru2017reinforced}
Pasunuru, R., Bansal, M.: Reinforced video captioning with entailment rewards.
  arXiv preprint arXiv:1708.02300  (2017)

\bibitem{paulus2017deep}
Paulus, R., Xiong, C., Socher, R.: A deep reinforced model for abstractive
  summarization. arXiv preprint arXiv:1705.04304  (2017)

\bibitem{phan2017consensus}
Phan, S., Henter, G.E., Miyao, Y., Satoh, S.: Consensus-based sequence training
  for video captioning. arXiv preprint arXiv:1712.09532  (2017)

\bibitem{ranzato2015sequence}
Ranzato, M., Chopra, S., Auli, M., Zaremba, W.: Sequence level training with
  recurrent neural networks. arXiv preprint arXiv:1511.06732  (2015)

\bibitem{rennie2016self}
Rennie, S.J., Marcheret, E., Mroueh, Y., Ross, J., Goel, V.: Self-critical
  sequence training for image captioning. arXiv preprint arXiv:1612.00563
  (2016)

\bibitem{DBLP:journals/corr/RiedlY14}
Riedl, M.O., Young, R.M.: Narrative planning: Balancing plot and character.
  CoRR  \textbf{abs/1401.3841} (2014), \url{http://arxiv.org/abs/1401.3841}

\bibitem{rush2015neural}
Rush, A.M., Chopra, S., Weston, J.: A neural attention model for abstractive
  sentence summarization. arXiv preprint arXiv:1509.00685  (2015)

\bibitem{see2017get}
See, A., Liu, P.J., Manning, C.D.: Get to the point: Summarization with
  pointer-generator networks. arXiv preprint arXiv:1704.04368  (2017)

\bibitem{sharma2017relevance}
Sharma, S., Asri, L.E., Schulz, H., Zumer, J.: Relevance of unsupervised
  metrics in task-oriented dialogue for evaluating natural language generation.
  arXiv preprint arXiv:1706.09799  (2017)

\bibitem{Stede:1996:SRT:974680.974687}
Stede, M.: Scott r. turner, the creative process. a computer model of
  storytelling and creativity. hillsdale, nj: Lawrence erlbaum, 1994. isbn
  0-8058-1576-7, \&pound;49.95. 298 pp. Nat. Lang. Eng.  \textbf{2}(3),
  277--285 (Sep 1996), \url{http://dl.acm.org/citation.cfm?id=974680.974687}

\bibitem{sutskever2014sequence}
Sutskever, I., Vinyals, O., Le, Q.V.: Sequence to sequence learning with neural
  networks. In: Advances in neural information processing systems. pp.
  3104--3112 (2014)

\bibitem{tu2016modeling}
Tu, Z., Lu, Z., Liu, Y., Liu, X., Li, H.: Modeling coverage for neural machine
  translation. arXiv preprint arXiv:1601.04811  (2016)

\bibitem{vinyals2015pointer}
Vinyals, O., Fortunato, M., Jaitly, N.: Pointer networks. In: Advances in
  Neural Information Processing Systems. pp. 2692--2700 (2015)

\bibitem{wang2017video}
Wang, X., Chen, W., Wu, J., Wang, Y.F., Wang, W.Y.: Video captioning via
  hierarchical reinforcement learning. arXiv preprint arXiv:1711.11135  (2017)

\bibitem{williams1992simple}
Williams, R.J.: Simple statistical gradient-following algorithms for
  connectionist reinforcement learning. In: Reinforcement Learning, pp. 5--32.
  Springer (1992)

\bibitem{zhang2017sentence}
Zhang, X., Lapata, M.: Sentence simplification with deep reinforcement
  learning. arXiv preprint arXiv:1703.10931  (2017)

\end{thebibliography}

\end{document}